%% file: main.tex
\documentclass[11pt,a4paper]{article}
\usepackage{microtype}
\usepackage{times,latexsym}
\usepackage{url}
\usepackage[T1]{fontenc}
\usepackage{amsmath}
\usepackage{booktabs, tabularx, multirow, graphics, hhline, boldline}
\usepackage{array, amssymb}
\usepackage{enumitem}
\usepackage{graphicx, color}
\usepackage{cite}
\usepackage{xr}
\usepackage{hyperref}
\usepackage[normalem]{ulem}
\usepackage[table,xcdraw]{xcolor}
\usepackage{caption}
\usepackage{subcaption}
\usepackage{tablefootnote}

\usepackage[acceptedWithA]{tacl2021v1}

\usepackage{xspace,mfirstuc,tabulary}

\newif\iftaclinstructions
\taclinstructionsfalse 
\iftaclinstructions
\renewcommand{\confidential}{}
\renewcommand{\anonsubtext}{(No author info supplied here, for consistency with
TACL-submission anonymization requirements)}
\newcommand{\instr}
\fi

\iftaclpubformat 

\else

\fi

\newif\ifdraft
\draftfalse

\definecolor{dkgreen}{RGB}{0,179,36}
\definecolor{dkred}{RGB}{240,0,0}
\definecolor{dkblue}{RGB}{0,100,240}
\definecolor{dkorange}{RGB}{230,115,0}
\definecolor{dkbrown}{RGB}{170,50,0}
\definecolor{chestnut}{rgb}{0.8, 0.36, 0.36}
\definecolor{pink}{RGB}{255,0,247}
\definecolor{amber}{rgb}{1.0, 0.75, 0.0}
\definecolor{amethyst}{rgb}{0.6, 0.4, 0.8}

\ifdraft
\newcommand{\kledit}[1]{\textcolor{blue}{{#1}}}

\newcommand{\spsedit}[1]{\textcolor{amethyst}{{#1}}}

\newcommand{\apremove}[1]{\textcolor{dkorange}{{\sout{#1}}}}
\newcommand{\klremove}[1]{\textcolor{blue}{{\sout{#1}}}}
\newcommand{\spsremove}[1]{\textcolor{amethyst}{{\sout{#1}}}}

\newcommand{\todo}[1]{\textcolor{dkred}{[{\it ToDo: #1}]}}

\newcommand{\note}[1]{\textcolor{dkred}{[{\it Note: #1}]}}
\else
\newcommand{\kledit}[1]{\textcolor{black}{{#1}}}

\newcommand{\spsedit}[1]{\textcolor{black}{{#1}}}

\newcommand{\apremove}[1]{{}}
\newcommand{\klremove}[1]{{}}
\newcommand{\spsremove}[1]{{}}
\newcommand{\todo}[1]{}
\newcommand{\note}[1]{}
\fi

\newcommand{\fig}{Fig.}
\newcommand{\figs}{Figs.}
\newcommand{\tab}{Tab.}
\newcommand{\sect}{Sec.}
\newcommand{\sects}{Secs.}

\newcommand{\wavtovec}{{\it wav2vec2}}
\newcommand{\wavtovecB}{{\it wav2vec2-Base}}

\newcommand{\fastvgsp}{{\it  FaST-VGS+}}
\newcommand{\vghubert}{{\it  VG-HuBERT}}
\newcommand{\avhubert}{{\it AV-HuBERT}}
\newcommand{\hubert}{{\it HuBERT}}

\newcommand{\hubertB}{{\it HuBERT-Base}}

\newcommand{\wavlm}{{\it WavLM}}
\newcommand{\wavlmB}{{\it WavLM-Base}}
\newcommand{\wavlmL}{{\it WavLM-Large}}
\newcommand{\baseM}{{\it Base}}
\newcommand{\largeM}{{\it Large}}

\DeclareMathOperator*{\argmax}{arg\,max}

\title{What Do Self-Supervised Speech Models Know About Words?}

\author{
Ankita Pasad \quad Chung-Ming Chien \quad Shane Settle \quad Karen Livescu
\\
Toyota Technological Institute at Chicago
\\
\{ankitap,cmchien,settle.shane,klivescu\}@ttic.edu
}

\date{}

\begin{document}

\maketitle

\begin{abstract}
Many self-supervised speech models (S3Ms) have been introduced over the last few years, improving performance and data efficiency on various speech tasks. However, these empirical successes alone do not give a complete picture of what is learned during pre-training. Recent work has begun analyzing how S3Ms encode certain properties, such as phonetic and speaker information, but we still lack a proper understanding of knowledge encoded at the word level and beyond. In this work, we use lightweight analysis methods to study segment-level linguistic properties---word identity, boundaries, pronunciation, syntactic features, and semantic features---encoded in S3Ms. We present a comparative study of layer-wise representations from ten S3Ms and find that (i) the frame-level representations within each word segment are not all equally informative, and (ii) the pre-training objective and model size heavily influence the accessibility and distribution of linguistic information across layers. We also find that on several tasks---word discrimination, word segmentation, and semantic sentence similarity---S3Ms trained with visual grounding outperform their speech-only counterparts. Finally, our task-based analyses demonstrate improved performance on word segmentation and acoustic word discrimination while using simpler methods than prior work.\footnote{Codebase: \href{https://github.com/ankitapasad/layerwise-analysis}{https://github.com/ankitapasad/layerwise-analysis}}

\end{abstract}

\input{sections/intro}
\input{sections/related-work}

\input{sections/methods}

\input{sections/experiments}
\input{sections/findings-frame}

\input{sections/findings-span}
\input{sections/findings-domain}

\input{sections/conclusion} 
\input{sections/ack}

\bibliographystyle{acl_natbib}
{\small
\bibliography{refs}
}


\end{document}

%% file: sections/intro.tex
\section{Introduction}
Self-supervised speech models (S3Ms) are effective across a variety of applications~\cite{mohamed2022self, yang2021superb}, including lower-level tasks such as speaker identification~\cite{chen2022wavlm} and speech recognition~\cite{baevski2020wav2vec, hsu2021hubert} as well as more linguistically complex spoken language understanding tasks~\cite{shon2022slue, tsai2022superb, pasad2021use, wu2023wav2seq, ashihara2023speechglue}. However, downstream task performance alone does not reveal what knowledge is learned during pre-training and where it is encoded.

Recent work has begun studying the acoustic and phonetic content encoded in S3Ms~\cite{pasad2021layer, hsu2021hubert, ma2021probing, pasad2023comparative, abdullah2023information}, and the findings have in turn helped guide model development and use~\cite{pasad2021layer, feng2022silence, baevski2022data2vec, pasad2023comparative, van2021analyzing, liu2023self}. However, S3Ms may encode higher-level linguistic information as well, since their model architectures (typically based on self-attention~\cite{vaswani2017attention, gulati2020conformer}) use contextual information from the entire speech input. There has been little analysis thus far of the word-level information encoded in these models~\cite{sanabria2023analyzing, pasad2023comparative}. To fill this gap, our work addresses two key questions:  
(i) {\it how is word-related information distributed across frames within a word segment?} 
and (ii) {\it in which layers, and how well, do S3Ms encode segment-level pronunciation, syntactic, and semantic information?}

To investigate these questions, we use canonical correlation analysis, a standard lightweight analysis tool, along with unsupervised evaluations on tasks: acoustic word discrimination, word segmentation, and semantic sentence similarity. We present a comparative study across ten S3Ms differing in their pre-training objective, data modality, and model size. Some of the key findings include:
\begin{itemize}[leftmargin=*,noitemsep,nolistsep]
\item The form of the pre-training objective affects which intermediate layers correlate the most with word-level properties. (\sect~\ref{sec:res-cca}) 
\item Word-identifying information is concentrated close to the center of each 
segment. (\sect~\ref{sec:word-loc})
\item Pre-trained representations from different S3Ms require varying complexities of post-processing to use the encoded knowledge. (\sect~\ref{sec:info-access})
\item The visually grounded S3Ms are better than speech-only S3Ms on several tasks: word discrimination (\sect~\ref{sec:info-access}), segmentation (\sect~\ref{sec:word-seg}), and semantic similarity (\sect~\ref{sec:res-sts}). 
\item A task's evaluation domain impacts the relative ranking of S3Ms, but the individual layer-wise trends remain domain-invariant. (\sect~\ref{sec:findings-domain})
\item With a simple parameter-free word segmentation algorithm, S3M features outperform previous, more complex unsupervised approaches. (\sect~\ref{sec:word-seg})
\item With a single-layer RNN trained on a small amount of labeled data, S3M features achieve near-perfect word discrimination, outperforming prior work by a large margin. (\sect~\ref{sec:info-access})
\end{itemize}

%% file: sections/related-work.tex
\section{Related work}
\label{sec:related-word}
The research community has begun investigating how S3Ms encode a number of properties such as speaker identity~\cite{fan2020exploring, van2021analyzing, chen2022wavlm, feng2022silence, liu2023self}, para-linguistics~\cite{shah2021all, li2023exploration}, articulatory and prosodic features~\cite{ji2022predicting, kim2022automatic, banno2023proficiency}, and phones~\cite{pasad2021layer, hsu2021hubert, ma2021probing, pasad2023comparative, abdullah2023information}.
Work on generative models based on S3Ms 
also suggests that some S3Ms learn phone-like sub-word units~\cite{lakhotia2021generative, nguyen2023generative}.  

Analysis on higher-level units, such as words in S3Ms, has been limited. 
\citet{pasad2023comparative} analyzed the extent to which different layers of S3Ms encode word identity, and \citet{sanabria2023analyzing} found that pooled S3M representations over word segments perform competitively on the task of acoustic word discrimination.  These results indicate that S3Ms encode word-identifying information.  However, it is not clear from prior work how word information is distributed across frames and what aspects of words are encoded, such as their pronunciation, syntactic properties, or semantic properties; our work addresses this gap.

Task-specific probing classifiers have been a common analysis tool for speech models~\cite{belinkov2019analysis,palaskar2019learned,prasad2020accents} including S3Ms~\cite{hsu2021hubert,baevski2021unsupervised,ma2021probing,shah2021all,shen2023wave}. While these probes provide an intuitive evaluation measure, 
the design decisions involved in training task-specific classifiers have confounding effects, making the scores hard to interpret~\cite{hewitt2019designing, ravichander2020probing, belinkov2022probing}. We use canonical correlation analysis (CCA)~\cite{harold1936relations} and training-free task-based evaluation, thus bypassing the dependence on task-specific classifiers. 

CCA and its more robust variants have been previously used to compare representations within and across neural networks~\cite{raghu2017svcca, kornblith2019similarity}, to study text representation models~\cite{voita2019bottom,tsvetkov2016correlation,saphra2019understanding}, and more recently for the analysis of S3M representations~\cite{pasad2021layer, li2023exploration, pasad2023comparative, yang2023can}. While classifiers require discrete labels, CCA can be used with both discrete and continuous-valued labels. CCA is also computationally inexpensive and has a closed-form solution.

Word similarity (WordSim) tasks~\cite{faruqui2014community} have been commonly used for intrinsic evaluation of word vectors. However, previous work has observed some problematic aspects of these tasks~\cite{faruqui2016problems}, including that WordSim performance is not well-correlated with extrinsic evaluation, whereas CCA-based evaluation more consistently tracks downstream task performance~\cite{tsvetkov2016correlation}.
Our work also shares some motivation with the Zero Resource Speech Benchmark~\cite{nguyen2020zero}, but tasks in this benchmark require encoding isolated word segments and/or discretizing the representations. Our analyses use word segment representations extracted in context, in order to match the most common use cases of S3Ms.

The task of acoustic word discrimination (AWD)~\cite{carlin2011rapid} has been commonly used to evaluate segment-level acoustic word embeddings, using both unsupervised~\cite{levin2013fixed,peng2020correspondence,algayres2020evaluating,van2021comparison} and supervised models~\cite{kamper2016cnn_awe,settle2016discriminative,he2017multiview,algayres2020evaluating}.  To our knowledge, only~\citet{sanabria2023analyzing} and~\citet{van2021comparison} have studied the use of S3Ms for generating acoustic word embeddings for AWD.  \citet{van2021comparison} used "first-generation" S3Ms while~\cite{sanabria2023analyzing} analyzed two of the same modern S3Ms we study here.  Our work provides a more comprehensive study of using multiple approaches (unsupervised mean pooling, dynamic time warping, and supervised models using S3M features), as well as a new state-of-the-art for one of the most commonly used AWD benchmarks.

Word unit discovery and segmentation are common benchmark tasks that have also been used to study speech representations~\cite{dunbar2020zero, nguyen2022word, sanabria2021difficulty, algayres2022dp, bhati2021segmental, cuervo2022contrastive, ten2007computational}.
Previous work studying the segmentation capabilities of S3Ms includes~\cite{kamper2022word}, based on first discovering phone-like units and then using them to discover word segments, and~\cite{peng2022word}, based on thresholding the attention map of a visually grounded speech model (VG-HuBERT, which we also use here).  Our study complements this prior work by comparing the layer-wise performance of a large number of S3Ms for this task and showing that a simple, training-free segmentation algorithm performs very competitively.

Textual sentence similarity is a classic task in NLP~\cite{conneau2018senteval}, but there are only a few studies investigating {\it spoken} utterance representations for this task~\cite{merkx2021semantic, merkx2023modelling, zhu2022bootstrapping}. Some downstream tasks in the SUPERB benchmark~\cite{yang2021superb} successfully use spoken utterance representations from frozen S3Ms, represented by a single mean-pooled vector. We complement our understanding of the capabilities of pooled utterance representations by performing a broad cross-model comparison.

%% file: sections/methods.tex
\section{Analysis methods}
\label{sec:analysis}
We extract frame-level and span-level representations from all layers of each S3M. We use CCA to compare word segment representations with various linguistic vectors (\sect~\ref{sec:ling-prop}) and also investigate encoded properties using training-free approaches for several tasks: (i) acoustic word discrimination (\sect~\ref{sec:method-word-disc}), (ii) word segmentation (\sect~\ref{sec:method-word-seg}), and (iii) sentence-level semantic similarity (\sect~\ref{sec:method-sts}).

\subsection{CCA-based analysis} 
\label{sec:ling-prop}
CCA~\cite{harold1936relations} is a statistical technique that measures the relationship between two continuous-valued random vectors by evaluating the maximum correlations between their linear projections. 
 CCA takes as input $n$ pairs of vectors $\{(x_1, y_1), ..., (x_n, y_n)\}$, sampled from the random vectors (or ``views") $X\in\mathbb{R}^{d_1}, Y\in\mathbb{R}^{d_2}$, and returns {\it canonical correlations}, a correlation-based measure of similarity between the two views. First, we solve for the directions of maximum correlation between linear projections of $X$ and $Y$: $ v_1, w_1 = \argmax_{v, w} \text{corr}(v^TX, w^TY) $.
The subsequent directions $v_i, w_i \; \forall \ i \in [2, \min(d_1, d_2)]$ maximize the same correlation subject to each new projection being uncorrelated with others in the same view. This problem has a closed-form solution requiring one singular vector decomposition. 

We use projection-weighted CCA (PWCCA)~\cite{morcos2018insights}, a robust CCA variant commonly used in recent analysis studies~\cite{voita2019bottom, pasad2021layer, pasad2023comparative, yang2023can}. The value of a PWCCA score lies between 0 and 1. Specifically, we use PWCCA to compare word segment representations with various linguistic vectors (\tab~\ref{tab:cca-exp}):

\input{tables/cca-exp.tex}

{\bf Word identity.}
We measure how well S3M representations encode word identity by comparing them with word IDs. For CCA computation, we follow \citet{pasad2023comparative}'s approach and convert the discrete word IDs to one-hot vectors. We use this analysis to examine the location of word information within the word segment. 

{\bf Acoustically grounded word embeddings.}
Acoustically grounded word embeddings (AGWEs) are written word embeddings trained jointly with acoustic word embeddings (AWE), i.e., representations of spoken word segments~\cite{he2017multiview,settle2019acoustically,hu2020multilingual}. A contrastive learning objective jointly optimizes AWE and AGWE models such that AWE and AGWE of the same word are closer than for pairs of different words.
We use AGWEs obtained from joint AWE+AGWE training on the LibriSpeech corpus~\cite{panayotov2015librispeech}.\footnote{The AGWEs used here are trained similarly to~\cite{shi2021whole} and are made available by \citet{pasad2021layer}: \href{https://github.com/ankitapasad/layerwise-analysis}{https://github.com/ankitapasad/layerwise-analysis}}
We expect CCA similarity with AGWEs to measure word-level pronunciation information encoded by the S3Ms.

{\bf Syntactic features.}
\label{sec:ptb}
\citet{tsvetkov2016correlation} construct syntactic vectors from the Penn Treebank (PTB)~\cite{marcus1993building}. For each word, an empirical probability is calculated for each of the 45 part-of-speech (POS) tags based on frequencies in the tagged corpus. This results in 45-dimensional syntactic vectors and each vector sums to 1.
For example, ``light" has values 0.52, 0.41, 0.05, and 0.02 for noun ({\it NN}), adjective ({\it JJ}), proper noun ({\it NNP}), and verb ({\it VB}) attributes, respectively, and zero for the rest.

{\bf Semantic features.}
\label{sec:semcor-glove}
\citet{tsvetkov2015evaluation} exploit word sense annotations in SemCor~\cite{miller1993semantic}, a WordNet-annotated version of the Brown Corpus. For each word, an empirical probability is calculated for each sense attribute (26 nouns and 15 verbs), based on their frequencies in the labeled corpus. This results in 41-dimensional semantic vectors and each vector sums to 1.
For instance, the vector for the word ``family" has a value of 0.96 and 0.04 for {\it NN.GROUP} and {\it NN.ACT} attributes, respectively, and zero for the rest. 

The resulting embedding space puts words with similar attributes closer together. For instance, the semantic vector of ``family" is most similar to words with a high value for the {\it NN.GROUP} attribute: government, leaders, elite, platoon. This behavior differs from that of a more fine-grained distributional embedding space such as GloVe~\cite{pennington2014glove} where some of the nearest neighbors for ``family" are husband, father, mother, sister, and wife. 

We do not compare S3M representations with learned text embeddings (such as GloVe or BERT~\cite{devlin2018bert} representations). Although these embeddings possibly encode richer text representations than our linguistic features, they also contain a mix of syntactic and semantic information. This would not allow us to study the syntactic and semantic features separately.

\subsection{Acoustic word discrimination} 
\label{sec:method-word-disc}
Acoustic word discrimination (AWD) is the task of determining whether or not a pair of acoustic waveform segments $({\bf X}_i, {\bf X}_j)$ correspond to the same word~\cite{carlin2011rapid}. A measure of dissimilarity between 
${\bf X}_i$ and ${\bf X}_j$
is computed, and the pair 
is predicted to be ``the same" if their dissimilarity falls below a threshold and ``different" otherwise. AWD performance is reported as average precision, i.e., the area under the precision-recall curve generated by varying the threshold. 

We use S3Ms for AWD in three ways. 
{\it pool-AWD} compares cosine distance after mean-pooling the frame-level features.
{\it DTW-AWD} computes a dynamic time warping distance between segments using the cosine distance between their frame-level features.
{\it RNN-AWD} trains a recurrent neural network on the frame-level representations, following \citet{shi2021whole}'s approach but using phone sequences for supervision as in~\cite{hu2020multilingual}.
 
\subsection{Word segmentation} 
\label{sec:method-word-seg}
\begin{figure}[tb]
 \centering
 \centerline{\includegraphics[width=0.9\linewidth]{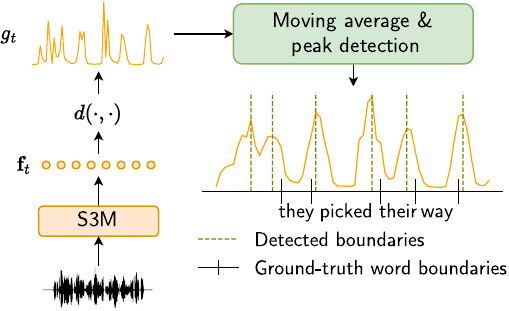}}
\caption{\it Our word segmentation algorithm.
}
  \label{fig:wordseg-illustration}
\vspace{-0.6cm}
\end{figure}

We ask how well S3M representations can perform word segmentation ``intrinsically''. We design a straightforward training-free algorithm to leverage the behavior of frame-level representations near word segment boundaries (see \fig~\ref{fig:wordseg-illustration}). 

Given a sentence comprising $T$ frames, first, we extract the frame-level features $\mathbf{f}_t$ ($1 \leq t \leq T$) from an S3M layer and perform mean and variance normalization, for each channel, across all $\mathbf{f}_t$'s to get the normalized $\hat{\mathbf{f}}_t$'s.
Then we compute the dissimilarity between adjacent frames $d(\cdot, \cdot)$ to get $g_t = d(\hat{\mathbf{f}}_{t+1}, \hat{\mathbf{f}}_t)$, and smooth $g_t$ with a moving average.
Finally, we use a peak detection algorithm to identify adjacent frames with higher dissimilarity than the surrounding frames. 
While peak detection algorithms have been commonly used for phoneme ~\cite{kreuk2020self, cuervo2022contrastive, rasanen2011blind} and word segmentation~\cite{bhati2021segmental, cuervo2022contrastive}, most prior word segmentation methods (with the exception of~\cite{peng2022word}) have relied on explicit training of the segmentation models while our approach does not.

The detected word boundaries are evaluated using standard metrics:
precision, recall, F1-score, and R-value, using a tolerance window of 20ms following prior work~\cite{kamper2022word}.

\subsection{Sentence-level semantic similarity}
\label{sec:method-sts}
Finally, we ask whether S3Ms encode any semantic content at the utterance level. We evaluate utterance representations from S3Ms on {\it spoken STS}~\cite{merkx2021semantic}, a spoken (read) version of the popular semantic textual similarity (STS) dataset~\cite{conneau2018senteval}. STS consists of sentence pairs annotated with a semantic similarity judgment. For each utterance in a pair, we extract a sentence-level representation from an S3M layer and use cosine similarity between these representations to predict semantic similarity. 
We report Spearman's $\rho$ correlation between the annotated human judgments and the predicted similarity scores.

%% file: tables/cca-exp.tex
\begin{table}[tb]
\centering
\small
\resizebox{\columnwidth}{!}{%
\begin{tabular}{ll}
\hlineB{2}
\begin{tabular}[c]{@{}l@{}}Linguistic property\end{tabular} & Attribute vector (dimension) \\\hlineB{2}
\begin{tabular}[c]{@{}l@{}}word identity\end{tabular}              & one-hot embeddings (500)                                     \\[12pt]
\begin{tabular}[c]{@{}l@{}}word pronunciation\end{tabular}                                                                
& \begin{tabular}[c]{@{}l@{}}acoustically grounded\\ word embeddings (128)\end{tabular}                                                           \\[12pt]
\begin{tabular}[c]{@{}l@{}}part-of-speech tags\end{tabular}                                                            & \begin{tabular}[c]{@{}l@{}}Attributes derived from\\ PTB$^1$ (45)\end{tabular}                                                       \\[12pt]
\begin{tabular}[c]{@{}l@{}}semantic attributes\end{tabular}                                            & \begin{tabular}[c]{@{}l@{}}Attributes derived from\\ SemCor\tablefootnote{\href{https://github.com/ytsvetko/qvec/tree/master/oracles}{https://github.com/ytsvetko/qvec/tree/master/oracles}} (41)\end{tabular}                                                            
\\\hlineB{2}
\end{tabular}
}
\caption{\it Linguistic properties that we compare to S3M representations via CCA.}
\vspace{-0.6cm}
\label{tab:cca-exp}
\end{table}

%% file: sections/experiments.tex
\section{Experiment details}
\label{sec:exp-details}
We present analysis for {\it ten S3Ms} differing in (i) pre-training objective, (ii) data modality (using either speech or image-speech pairs), and (iii) model size. The pre-trained checkpoints are obtained from publicly available sources.
\footnote{
\wavtovec: \href{https://github.com/facebookresearch/fairseq/tree/main/examples/wav2vec}{https://github.com/facebookresearch/fairseq} \\
\hubert: \href{https://github.com/facebookresearch/fairseq/tree/main/examples/hubert}{https://github.com/facebookresearch/fairseq} \\ 
\wavlm: \href{https://github.com/microsoft/unilm/tree/master/wavlm}{https://github.com/microsoft/unilm/tree/master/wavlm}
\avhubert: \href{https://github.com/facebookresearch/av_hubert}{https://github.com/facebookresearch/av\_hubert}\\
\fastvgsp: \href{https://github.com/jasonppy/FaST-VGS-Family}{https://github.com/jasonppy/FaST-VGS-Family}\\
\vghubert: \href{https://github.com/jasonppy/word-discovery}{https://github.com/jasonppy/word-discovery}
.}
For word-level analysis on LibriSpeech~\cite{panayotov2015librispeech} (\sects~\ref{sec:exp-cca} and~\ref{sec:exp-awd}), we use ground-truth word alignments generated by the Montreal Forced Aligner~\cite{mcauliffe2017montreal, lugosch2019speech}.

\subsection{Background on S3Ms}
\label{sec:background}
S3Ms are trained with an objective function formulated to solve a pretext task on unlabeled data. In a typical model architecture, the raw audio (or filter bank features) is first passed through convolutional layers (or a linear projection). Then, the resulting frame-level \emph{local} features are processed through self-attention layers. The models we use have 7 convolutional (or 1 linear) and 12 or 24 transformer layers.

All the models in this work use a masking-based pretext task, using both the left and the right context to recover the masked segment (target). The target comes from either the local features (wav2vec 2.0$^4$ (\wavtovec)~\cite{baevski2020wav2vec} and \fastvgsp~\cite{peng2022self}) or from one of the intermediate transformer layers, represented as a discrete cluster ID  (\hubert\footnote{\wavtovec \ and \hubert \ {\it Base} models are pre-trained on 960 hours LibriSpeech, and the corresponding {\it Large} models on 60k hours LibriLight data.}~\cite{hsu2021hubert}, \wavlm\footnote{\wavlm {\it -Base} is pre-trained on 960 hours LibriSpeech and \wavlm{\it -Large} on 94k hours consisting of LibriLight, GigaSpeech, and VoxPopuli.}~\cite{chen2022wavlm}, \avhubert\footnote{\avhubert \ models are pre-trained on LRS3.}~\cite{shi2022robust}, and \vghubert~\cite{peng2022word}). Models of the first type are trained with a contrastive loss and the latter with a classification loss. The classification loss for \wavlm \ uses cluster IDs from \hubert's intermediate layers (the same ones used in \hubert's iterative pre-training). Unlike \hubert, \wavlm \ augments the input data to simulate noisy/overlapped speech.

The visually grounded\footnote{It is arguable whether such visually grounded models are "self-supervised" since the visual signal provides a form of supervision. We include them here since they are in many ways similar to speech-only S3Ms and have similar use cases.} \fastvgsp \ and \vghubert \ models are initialized with pretrained \wavtovecB\footnote{For \fastvgsp, CNN, self-attention, and cross-attention layers are added before training on SpokenCOCO.} and \hubertB,\footnote{For \vghubert, the top 3 layers are reinitialized before training on SpokenCOCO.} \ respectively, thus providing a way to isolate and analyze the effect of visual grounding. These models are trained with a cross-modal contrastive loss with an appended {\it CLS} token, a fixed-dimensional utterance-level representation. \avhubert \ is trained on a lipreading dataset with a pre-training objective that uses multi-modal discrete units.
In the case of the audio-visual models, we use only the audio branch for our analysis. 

\subsection{CCA evaluation}
\label{sec:exp-cca}
For CCA similarity with word ID ({\it CCA-word}), we sample $\sim$7k word instances across 500 distinct words from the dev-clean subset of LibriSpeech. 
We represent the word segment using either a single frame, by mean-pooling across a quarter of the contiguous frames, or by mean-pooling across all the frames within the word boundaries. The single frame is sampled from one of five equidistant locations starting at the first frame. The quarter chunk of contiguous frames is extracted from one of the four quarters of the word segment. 

In all other CCA experiments, we obtain word segment representations by mean-pooling across all frames within the word boundaries and compare these representations with external linguistic embedding vectors (\tab~\ref{tab:cca-exp}). We sample 364k word instances across 9.9k distinct words from the LibriSpeech train-clean and train-other subsets. This sample includes the 8.6k and 4k word vocabularies from {\it PTB} and {\it SemCor} vectors, respectively.

We evaluate PWCCA with multiple data splits to avoid overfitting, using the implementation from~\citet{pasad2023comparative}.\footnote{\href{https://github.com/ankitapasad/layerwise-analysis}{https://github.com/ankitapasad/layerwise-analysis}} Specifically, we run each experiment with 5 train-val-test splits. In the result figures below, we plot the mean of the 5 runs along with a shadow around the mean corresponding to the minimum and maximum values.  For most result plots, the shading is not visible as there is negligible deviation among results across runs.

\subsection{Acoustic word discrimination}
\label{sec:exp-awd}
We evaluate {\it pool-AWD} and {\it DTW-AWD} on ``clean" and ``other" partitions of the LibriSpeech development set. 
Our {\it RNN-AWD} models are trained and evaluated on Switchboard~\cite{godfrey1992switchboard} data using the same train-dev-test split as prior work~\cite{carlin2011rapid,jansen2013weak_topdown,kamper2016cnn_awe,he2017multiview} with each partition containing approximately $10$k spoken word segments. We evaluate {\it pool-AWD} on our Switchboard dev set first to find the best layer before supervised {\it RNN-AWD} training. 
In all cases, spoken word segments are $0.5$-$2$s in duration, and segments used for evaluation on LibriSpeech and Switchboard span 5k and 3k word vocabularies, respectively. 

\subsection{Word segmentation}
\label{sec:exp-wseg}
We consider two measures of dissimilarity between neighboring frames, Euclidean distance and cosine distance.  We use a prominence-based algorithm~\cite{virtanen2020scipy} to detect peaks in the dissimilarity curve with a prominence value exceeding a specified threshold.
For each layer in each S3M, we grid search over the choice of distance metric, prominence value threshold, and moving-average window size. We choose the best combination based on F1-scores for word boundary detection on a randomly sampled subset of the LibriSpeech dev-clean split ($\sim$2k utterances).
We also evaluate all layers on the  Buckeye~\cite{pitt2005the} validation set, and the best layer of each S3M is evaluated on the Buckeye test set. 

\subsection{Sentence-level semantic similarity}
The natural speech recordings in {\it Spoken STS} constitute 5\% (638 sentence pairs) of the original STS corpus~\cite{merkx2021semantic}. Sentences in each pair are read by four speakers, and thus, each pair has 16 speaker combinations. Each spoken sentence is represented by mean-pooling all frame-level representations from an S3M layer. For \vghubert, we extract the utterance-level {\it CLS} token representation as well. As in previous work~\cite{merkx2021semantic, zhu2022bootstrapping}, the predicted score for each sentence pair is the mean of the cosine similarities between their representations for all speaker combinations.

%% file: sections/findings-frame.tex
\input{figs/res1-all}
\input{figs/res-dtw}
\section{Findings}
\label{sec:findings}
We present our findings in two parts. \sect~\ref{sec:frame-analysis} investigates the spoken word knowledge learned by S3Ms, and how this information is distributed across the frames of each word segment.\footnote{Model trends are consistent between LibriSpeech dev-clean and dev-other results, so only dev-clean results are shown.}
\sect~\ref{sec:findings2} looks at specific linguistic properties---pronunciation, syntactic, and semantic---of word-level and sentence-level representations.

\subsection{Analysis of frame-level representations}
\label{sec:frame-analysis}
We investigate the word-related information encoded by S3M layers in different frames across the word segment, specifically knowledge of word identity and word boundaries.

\subsubsection{Ease of accessing encoded information}
\label{sec:info-access}
\fig~\ref{fig:cca-word} shows layer-wise correlation scores with word ID vectors for all models.\footnote{Results from all models except \vghubert \ are replicated from \citet{pasad2023comparative}.} We investigate whether this word-identifying information, as evidenced by high CCA scores, is easily accessible by evaluating the word representations on AWD.

For {\it pool-AWD}, we see that the best and worst performing models have a large difference (\figs~\ref{fig:awe-clean},~\ref{fig:awe-swbd}) despite being similarly well-correlated with word ID vectors (\fig~\ref{fig:cca-word}). Next, we evaluate {\it DTW-AWD} on a subset of models (\fig~\ref{fig:dtw-clean}) and find that (i) all models have \kledit{a }better performance than {\it pool-AWD} with a reduced cross-model performance gap, and (ii) the cross-model ranking is also consistent with {\it pool-AWD}. The cross-model performance gap is further reduced with our supervised {\it RNN-AWD} experiments (\tab~\ref{tab:acoustic_word_discrimination}) and the cross-model ranking is consistent with the corresponding {\it pool-AWD} trends on Switchboard (\fig~\ref{fig:awe-swbd}). Our multi-view {\it RNN-AWD} model attains a near-perfect average precision, significantly outperforming previous work (by >10\% absolute).

These experiments suggest that some models (such as \wavtovec) distribute discriminative word information across frames in a way that is not easily extracted through mean-pooling and compared via cosine distance, indicating that more structured reasoning over the whole segment may be helpful, such as frame-level processing in our DTW and RNN experiments. 
A similar observation is made in prior work~\cite{sanabria2023analyzing} where representing words using sub-sampled and concatenated frames instead of mean-pooling gives the most relative improvement for \wavtovec.

We include a detailed discussion on the layer-wise trends and effect of evaluation domain (\figs~\ref{fig:awe-clean},~\ref{fig:awe-swbd}) in \sect~\ref{sec:findings-domain-awd}.

\input{tables/acoustic_word_discrimination}

\subsubsection{Are all frames equally informative?}
\label{sec:word-loc}

Next, we analyse frame-level representations to understand how word-identifying information is distributed within 
word boundaries.
We represent 
word segments either using individual frames at different locations or by pooling over frames spanning different quarters of the segment.

\begin{figure}[tbh]
 \centering
 \includegraphics[width=\linewidth, trim=0 270 0 0, clip]{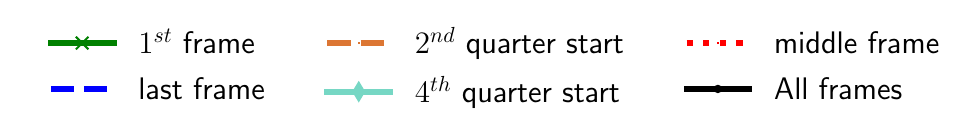}
 \includegraphics[width=\linewidth, trim=0 147 0 0, clip]{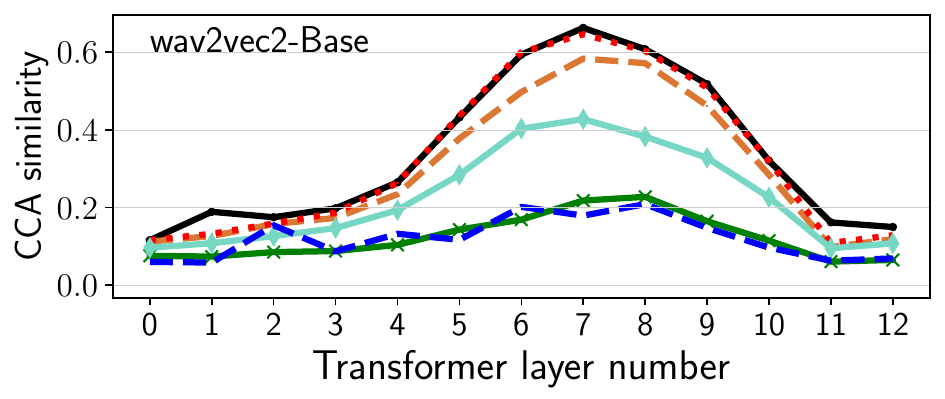}
 \caption{{\it Correlation with word identity for \wavtovecB \ when using a single frame to represent a word segment.}
 }
 \label{fig:word-loc-single}
\end{figure}

\begin{figure}[tbh]
 \centering
 \includegraphics[width=\linewidth, trim=0 260 0 0, clip]{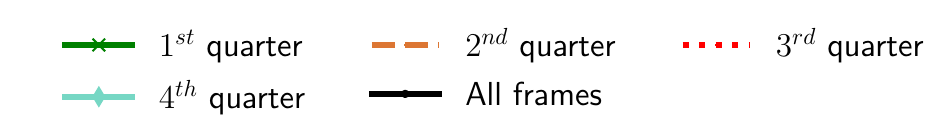}
 \includegraphics[width=\linewidth, trim=0 185 0 0, clip]{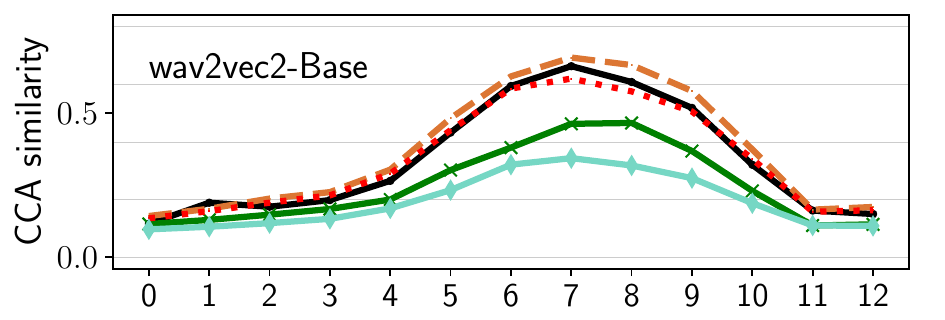}
 \includegraphics[width=\linewidth, trim=0 185 0 0, clip]{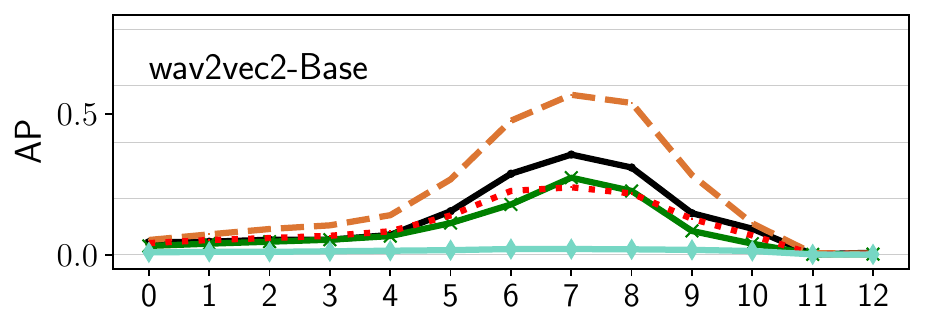}
 \includegraphics[width=\linewidth, trim=0 170 0 0, clip]{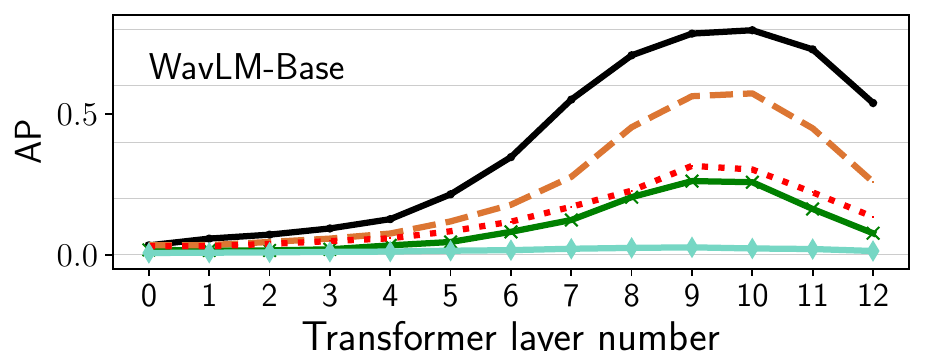}
 \caption{\it Correlation with word identity and AWD scores when pooling over segment quarters.}
 \label{fig:word-loc-quarter}
\end{figure}

We measure CCA scores between word segment representations and word ID and find that frames near the center of the word segment are most informative of the word identity (\figs~\ref{fig:word-loc-single} and \ref{fig:word-loc-quarter}). 
Specifically, the single center frame and the 2$^{nd}$ and 3$^{rd}$ quarter spans are all as highly correlated with the word identity as the mean-pooled representations. 
These findings are consistent across all S3Ms analyzed.

We see a similar word localization trend in the AWD evaluations (\fig~\ref{fig:word-loc-quarter}), but with a stronger bias toward the start of the segment. 
In particular, using the $2^{nd}$ quarter span alone yields a better AP 
for \wavtovecB \ and \hubertB. 
This gives a possible explanation for their better relative performance on {\it DTW-AWD} and {\it RNN-AWD} compared to {\it pool-AWD} since those approaches can adjust their focus to only the most relevant frames.

\subsubsection{Word segmentation}
\label{sec:word-seg}
\fig~\ref{fig:wordseg} shows the F1-scores of S3Ms on the word segmentation task.
All of the models demonstrate non-trivial word segmentation capability.\footnote{\avhubert~is not included in this experiment as its frame rate is 40 ms, which is larger than the maximum acceptable error of 20 ms on the Buckeye word segmentation task.}
We observe that visually grounded models consistently outperform their speech-only counterparts, possibly because of the visual context.
Further strengthening this hypothesis, we note that \vghubert \ has a minimal performance drop at the final few layers, unlike other S3Ms, which can be attributed to the proximity to the cross-modal loss. \fastvgsp \ does not see the same trend, while it is designed such that the final few layers we analyze here are only trained on the self-supervised loss but not the cross-modal loss.  
Similar unit discovery capabilities of visually grounded models have also been studied in prior work~\cite{harwath-glass-2017-learning, peng2022word}.
We include more discussion of performance comparison across S3Ms and across evaluation domain (\figs~\ref{fig:wseg-libri},~\ref{fig:wseg-buckeye}) in \sect~\ref{sec:findings-domain-wseg}.

\input{figs/res-wseg}

In \tab~\ref{tab:word_seg}, we compare the best-performing layers in our experiments with previous word segmentation algorithms that take S3M features as inputs. With our simple training-free method, we obtain the best F1 score of 41.0\% using the 10$^{th}$ layer of \vghubert. This outperforms a previously published attention-based approach using \vghubert~\cite{peng2022word} and a recent dynamic programming-based approach that also trains an autoencoder on top~\cite{kamper2022word}.
However, we note that our approach falls short in terms of R-values, implying that it tends to over-segment.
This can possibly be improved by designing different criteria for hyper-parameter selection, as our criterion is solely based on the F1 score. 

\addtocounter{footnote}{1}
\footnotetext{
GradSeg~\cite{fuchs2023unsupervised} also shows impressive word segmentation results on the Buckeye dataset, but they provide results only on the validation set, making it difficult to compare.
}
\addtocounter{footnote}{-1}

\input{tables/segmentation}

%% file: figs/res1-all.tex
\begin{figure*}[t]
\begin{minipage}[b]{1.0\linewidth}
\small

 \centering
 \centerline{\includegraphics[width=0.95\linewidth, trim=0 320 0 0, clip]{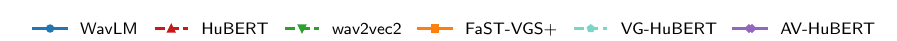}}
\end{minipage}
\begin{minipage}[b]{1.0\linewidth}

\centering
     \begin{subfigure}[b]{\linewidth}
         \centering
         \includegraphics[width=\linewidth, trim=0 235 0 0, clip]{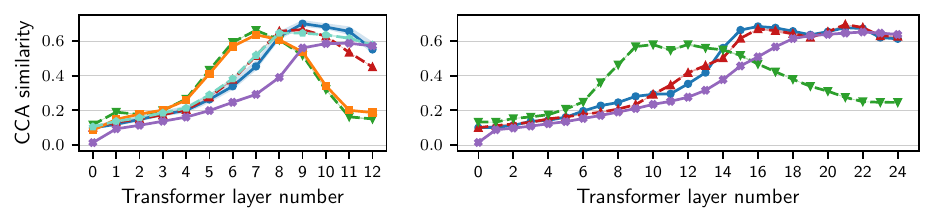}
         \caption{{\it CCA similarity between mean-pooled S3M word segment representations and word ID one-hot vectors.}
         }
         \label{fig:cca-word}
     \end{subfigure}
     
     \begin{subfigure}[b]{\linewidth}
         \centering
         \includegraphics[width=\linewidth, trim=0 235 0 0, clip]{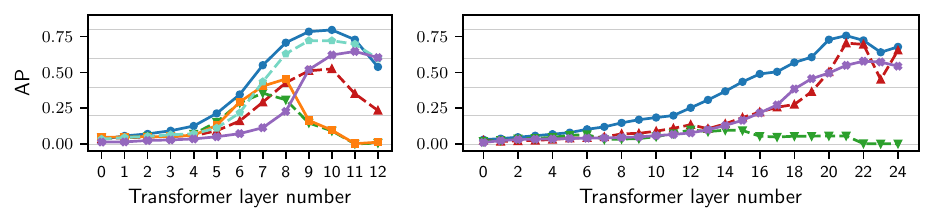}
         \caption{{\it pool-AWD scores on LibriSpeech dev-clean.}}
         \label{fig:awe-clean}
     \end{subfigure}

     \begin{subfigure}[b]{\linewidth}
         \centering
         \includegraphics[width=\linewidth, trim=0 235 0 0, clip]{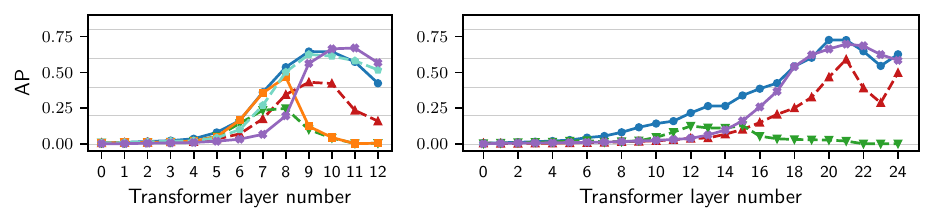}
         \caption{{\it pool-AWD scores on Switchboard dev.}}
         \label{fig:awe-swbd}
     \end{subfigure}
     
\end{minipage}
\vspace{-0.7cm}
\caption{\it Evaluation of the word-identifying information in mean-pooled word segment representations from \baseM \ (left) and \largeM \ (right) S3Ms.}

  \label{fig:scores1}
\vspace{-0.3cm}
\end{figure*}

%% file: figs/res-dtw.tex
\begin{figure}[tbh]
 \centering
 \centerline{\includegraphics[width=0.85\linewidth, trim=0 320 230 0, clip]{images/legend-scores.pdf}}
 \includegraphics[width=0.9\linewidth, trim=0 235 260 0, clip]{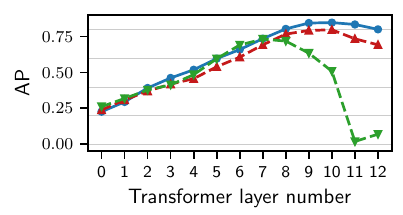}
 \caption{{\it DTW-AWD} results on LibriSpeech dev-clean.}
 \vspace{-0.1in}
 \label{fig:dtw-clean}
\end{figure}

%% file: tables/acoustic_word_discrimination.tex
\begin{table}[tbh]
  \centering
  \small
  \begin{tabular}{lc}
    \toprule
    Method      &   AP             \\
    \midrule
    Multi-View RNN~\cite{he2017multiview}\\ 
    $\,\,$ w/ log-Mel filterbank features  & 0.84\\
    \midrule
    $\,\,$ w/ \wavtovecB \ (L8)              & 0.93\\
    $\,\,$ w/ \hubertB \ (L9)                & 0.94\\
    $\,\,$ w/ \wavlmB \ (L10)                & 0.95\\
    $\,\,$ w/ \wavlmL \ (L20)                & \textbf{0.98}\\
    \bottomrule
  \end{tabular}
\caption{{\it RNN-AWD} performance on the Switchboard word discrimination test set~\cite{carlin2011rapid}; layer used specified in parentheses. 
}
\label{tab:acoustic_word_discrimination}
\end{table}

%% file: figs/res-wseg.tex
\begin{figure*}[tbh]
\begin{minipage}[b]{1.0\linewidth}
\small

 \centering
 \centerline{\includegraphics[width=0.95\linewidth, trim=0 320 0 0, clip]{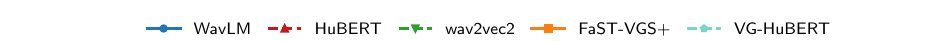}}
\end{minipage}
\begin{minipage}[b]{1.0\linewidth}

\centering
     \begin{subfigure}[b]{\linewidth}
         \centering
         \includegraphics[width=\linewidth, trim=0 235 0 0, clip]{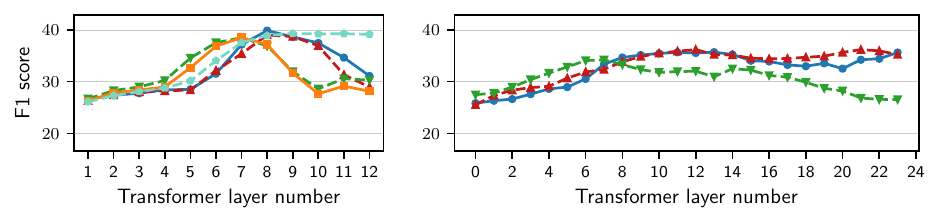}
         \caption{\it F1-scores on the LibriSpeech dev set}
         \label{fig:wseg-libri}
     \end{subfigure}
     \begin{subfigure}[b]{\linewidth}
         \centering
         \includegraphics[width=\linewidth, trim=0 235 0 0, clip]{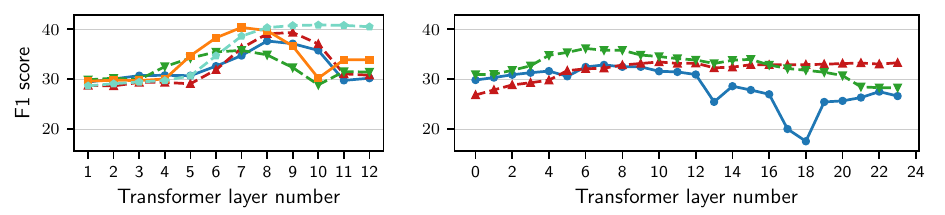}
         \caption{\it F1-scores on the Buckeye validation set}
         \label{fig:wseg-buckeye}
     \end{subfigure}
\end{minipage}
\vspace{-0.7cm}
\caption{{\it Unsupervised word segmentation using representations from \baseM \ (left) and \largeM \ (right) S3Ms}. }
  \label{fig:wordseg}
\vspace{-0.3cm}
\end{figure*}

%% file: tables/segmentation.tex
\begin{table}[tbh]
    \setlength{\tabcolsep}{2pt}
    \small
    \centering
    \begin{tabular}{l r r r r}
        \toprule
        Method & Prec. & Rec. & F1 & R-val. \\
        \midrule
        \textit{Prior work\protect\footnotemark} \\
        \hspace{2mm} DPDP~\cite{kamper2022word}  & 35.3 & 37.7 & 36.4 & 44.3 \\
        \hspace{2mm} \vghubert & \multirow{2}{*}{\bf{36.2}} & \multirow{2}{*}{32.2} & \multirow{2}{*}{34.1} & \multirow{2}{*}{\bf{45.6}} \\
        \hspace{2mm} \citep{peng2022word} & & & &\\
        \midrule
        \multicolumn{5}{l}{\textit{Ours (Best Layer)}} \\
        \hspace{2mm} \wavlmB \ (L8) &  31.9 & 45.7 & 37.6 &30.7\\ 
        \hspace{2mm} \hubertB \ (L9) & 33.8 &  46.6 & 39.2 & 34.9 \\ 
        \hspace{2mm} \wavtovecB \ (L7) & 27.0 & 47.2 & 34.3 & 8.9\\ 
        \hspace{2mm} \vghubert \ (L10) & 36.0 & \bf{47.6} & \bf{41.0} & 39.5\\ 
        \bottomrule 
    \end{tabular}
    \caption{{\it Word segmentation performance on the Buckeye test set. Higher is better for all metrics.}}
    \label{tab:word_seg}
\end{table}

%% file: sections/findings-span.tex
\subsection{Analysis of pooled span representations}
\label{sec:findings2}
\input{figs/res2-all}
Next, we measure how correlated word segment representations are with the other linguistic properties from \tab~\ref{tab:cca-exp}: pronunciation, syntactic (POS) attributes, and semantic attributes (\sect~\ref{sec:res-cca}), and \spsedit{we }evaluate mean-pooled utterance representations on sentence similarity (\sect~\ref{sec:res-sts}).
Our remaining word-level experiments consider mean-pooled word segment representations as these consistently correlate well with word ID (\figs~\ref{fig:word-loc-single} and \ref{fig:word-loc-quarter}). 

\subsubsection{Similarity with linguistic properties}
\label{sec:res-cca}
In \fig~\ref{fig:cca-scores}, we observe that models trained to recover local features (\wavtovec \ and \fastvgsp) have the highest correlation at central layers, specifically layers 5-7 for \baseM \ models and layers 8-11 for the \largeM \ model. The rest of the models are trained to recover discrete units from an intermediate layer and have the highest correlation at much higher layers. This dependence on the form of pre-training objective has been observed before for lower-level acoustic and phonetic features~\cite{pasad2023comparative}.

As seen for our other experiments (\figs~\ref{fig:scores1},~\ref{fig:wordseg}), the audio-visual models (\avhubert \ and \vghubert) see the least drop off in the final layers. These models are optimized with an audio-visual objective, suggesting that meaningful linguistic content is retained better with visual grounding. 

For all S3Ms, pronunciation content (\fig~\ref{fig:res-agwe}) is best correlated at lower layers than syntactic (\fig~\ref{fig:res-ptb}) and semantic properties (\fig~\ref{fig:res-semcor}). In \baseM \ models, the same set of intermediate layers is best correlated with both syntactic and semantic attributes. The \largeM \ models, on the other hand, have a more pronounced peak for semantic than syntactic content, which in turn has a narrower plateau than the word pronunciation trends (\fig~\ref{fig:cca-scores} {\it right}). 

This differs from some observations made for BERT, a pre-trained {\it text} model, where different linguistic features---such as POS, constituents, dependencies, and entities---are encoded best at different layers~\cite{tenney2019bert}. This difference is possibly because the speech pre-training objective is mostly local with much of the model capacity (i.e. the majority of the layers) devoted to inferring local acoustic and lower-level phonetic features. Meanwhile, text models that start with higher-level segmented sub-word units have the capacity to encode fine-grained linguistic properties in different layers. 
BERT's superiority in linguistic knowledge is supported by \citet{shen2023wave} where BERT outperforms \wavtovec \ and \hubert \ by 20\% relative on a parsing-related probing task.

\input{figs/res-spoken-sts}

\begin{figure}[ht]
 \centering
 \centerline{\includegraphics[width=8cm]{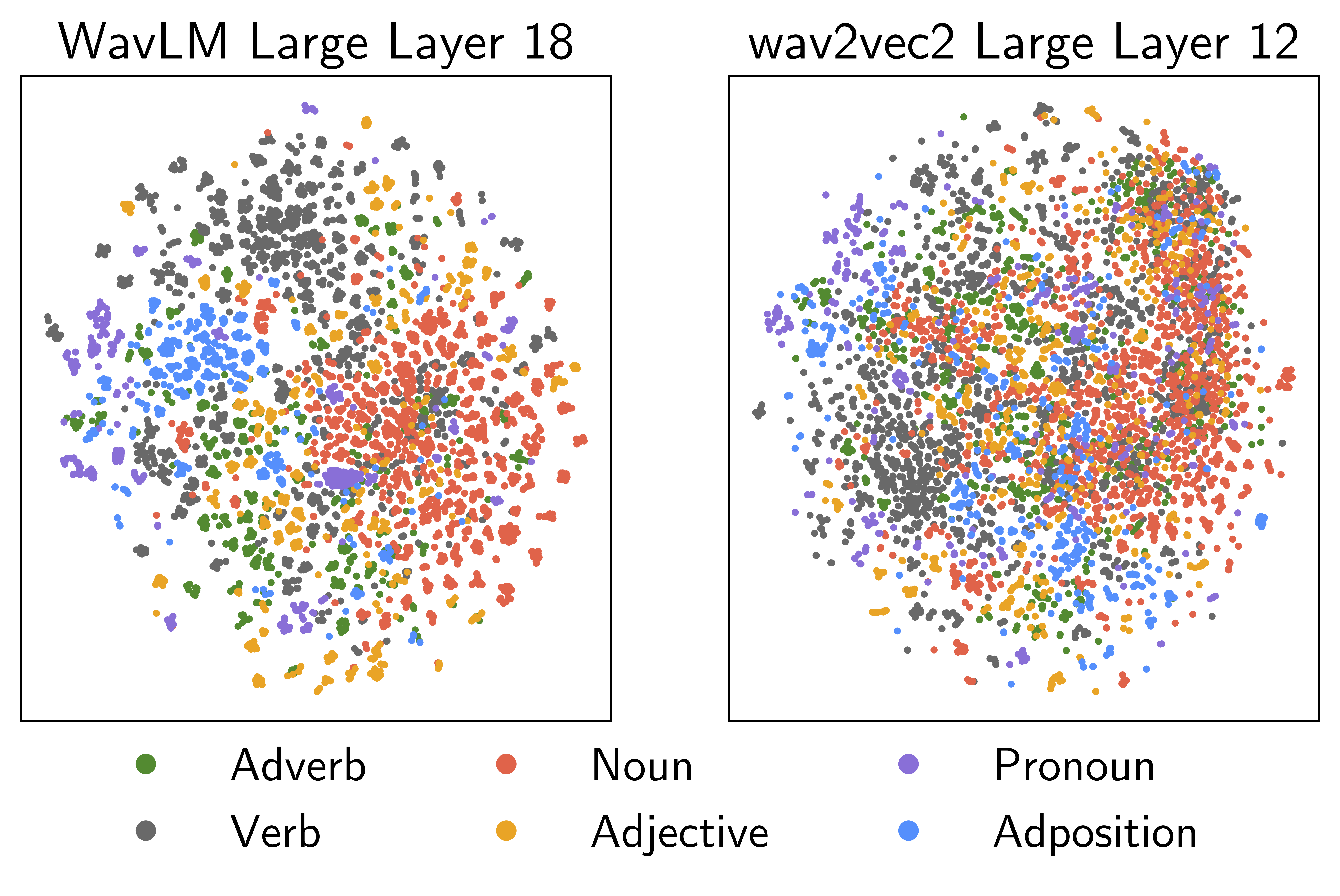}}
\vspace{-0.3cm}
\caption{\it Visualization of the embedding spaces of the intermediate layers of S3Ms. Each point represents one word sample. Only the 6 most common POS tags are shown.
}
  \label{fig:tsne-pos}
\vspace{-0.3cm}
\end{figure}

To qualitatively study the syntactic information encoded in S3M representations, we visualize the mean-pooled word representations from the layers with high correlation with the PTB syntactic vectors (\fig~\ref{fig:res-ptb}). We sample $\sim$7k word instances across 500 distinct words and apply t-SNE to project the word representations to 2 dimensions (\fig~\ref{fig:tsne-pos}).
We find that, for \wavlm, word samples with the same POS tag (especially for verbs, nouns, and adpositions) are encoded into vectors close to each other. However, the representations of \wavtovec\ are not as well-separated. These visualizations further corroborate our findings from CCA trends (\fig~\ref{fig:res-ptb}), where \wavlm\ shows a greater correlation than \wavtovec.

\subsubsection{Sentence-level semantics}
\label{sec:res-sts}
\fig~\ref{fig:sts} shows the layer-wise performance on the spoken sentence similarity task. We include two baselines: (i) {\it FBank} uses mean-pooled filter-bank features as a sentence representation, and (ii) {\it naive} text baseline reports the fraction of word overlap in text transcripts between a pair of sentences. Although the {\it naive} text baseline has a non-trivial correlation score of $0.4$, the best-performing layers outperform the baselines by at least 50\%. These results suggest that the mean-pooled S3M representations encode meaningful content beyond just the local acoustics and word identities.

The CLS token of \vghubert \ has the best correlation score of 0.64 at layer 11, closely followed by layer 8 of \vghubert \ and \fastvgsp, both visually grounded models. The speech-only S3Ms we analyze outperform other S3Ms previously evaluated on this task~\cite{merkx2021semantic, zhu2022bootstrapping}.\footnote{The comparison with \cite{merkx2021semantic} is based on Pearson's correlation, not reported here.} However, they all underperform a text oracle baseline from \citet{zhu2022bootstrapping} using self-supervised text embeddings ({\it SimCSE-unsup-RoBERTa}), which has a correlation score of 0.77.

%% file: figs/res2-all.tex
\begin{figure*}[t]
\begin{minipage}[b]{1.0\linewidth}
\small

 \centering
 \centerline{\includegraphics[width=0.95\linewidth, trim=0 318 0 0, clip]{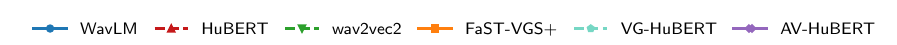}}
\end{minipage}
\begin{minipage}[b]{1.0\linewidth}

\centering     
     \begin{subfigure}[b]{\linewidth}
         \centering
         \includegraphics[width=\linewidth, trim=0 235 0 0, clip]{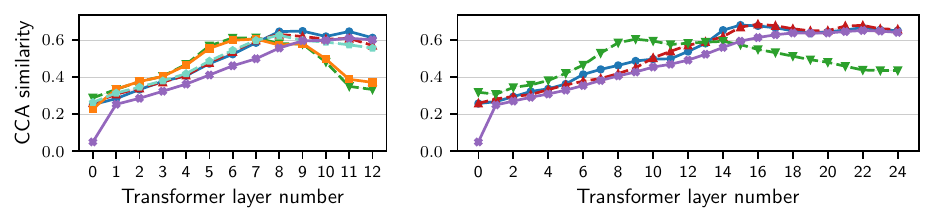}
         \caption{{\it Word pronunciation content; CCA similarity between S3M word segment representations and AGWEs.}}
         \label{fig:res-agwe}
     \end{subfigure}
 
     \begin{subfigure}[b]{\linewidth}
         \centering
         \includegraphics[width=\linewidth, trim=0 235 0 0, clip]{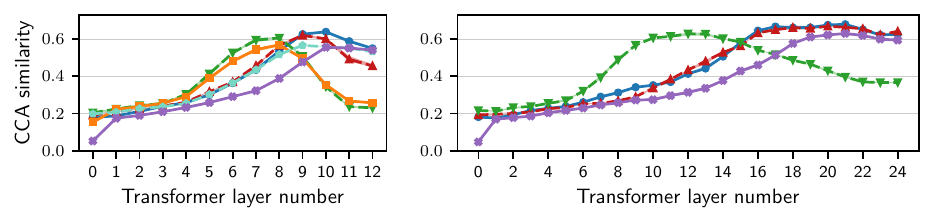}
         \caption{{\it Syntactic content; CCA similarity between S3M word segment representations and POS attributes.}}
         \label{fig:res-ptb}
     \end{subfigure}
     
     \begin{subfigure}[b]{\linewidth}
         \centering
         \includegraphics[width=\linewidth, trim=0 235 0 0, clip]{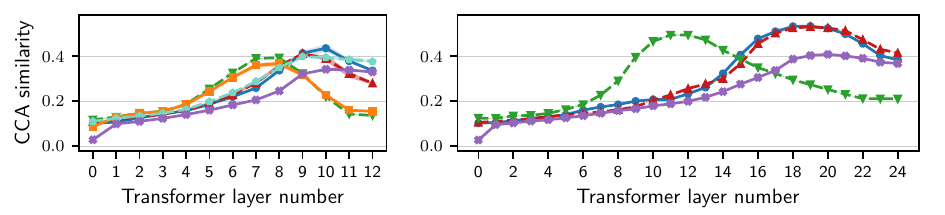}
         \caption{{\it Semantic content; CCA similarity between S3M word segment representations and SemCor attributes.}}
         \label{fig:res-semcor}
     \end{subfigure}
\end{minipage}
\vspace{-0.7cm}
\caption{\it Measure of different linguistic properties using CCA for \baseM \ (left) and \largeM \ (right) S3Ms}.
\label{fig:cca-scores}
\vspace{-0.7cm}
\end{figure*}

%% file: figs/res-spoken-sts.tex
\begin{figure*}[tbh]
 \centering
 \centerline{\includegraphics[width=0.95\linewidth, trim=0 310 0 0, clip]{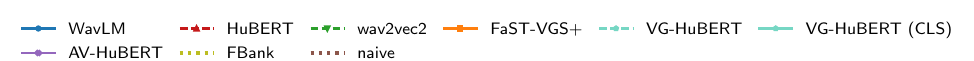}}
     \includegraphics[width=\linewidth, trim=0 235 0 0, clip]{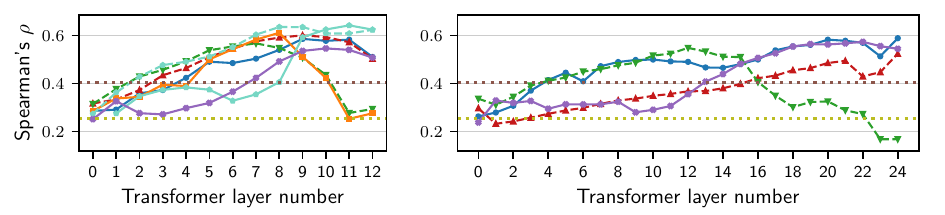}
\caption{\it Performance on spoken STS task using representations from \baseM \ (left) and \largeM \ (right) S3Ms. 
}
 \label{fig:sts}
\end{figure*}

%% file: sections/findings-domain.tex
\subsection{Effect of domain on task-based evaluation}
\label{sec:findings-domain}
Prior work evaluating S3Ms on downstream tasks has demonstrated how the relative ranking of S3Ms may be influenced by the domain of an S3M's pre-training data as well as the evaluation methodology~\cite{hsu2021robust, yang2021superb, tsai2022superb, zaiem2023speech}. For instance, similarly to all our task-based experiments (\figs~\ref{fig:awe-clean}, \ref{fig:wordseg}, and \ref{fig:sts}), the SUPERB benchmarks~\cite{yang2021superb, tsai2022superb}\footnote{\href{https://superbbenchmark.org/leaderboard}{https://superbbenchmark.org/leaderboard}} and \citet{zaiem2023speech} report instances where some \largeM \ S3Ms under-perform their \baseM \ counterparts on downstream tasks.

Next, we discuss our takeaways related to the effect of (mis-)match between the domain of pre-training data and task data on some of our analysis experiments.

\subsubsection{Acoustic word discrimintaion}
\label{sec:findings-domain-awd}
We evaluate {\it pool-AWD} on both LibriSpeech (\fig~\ref{fig:awe-clean}), a read speech domain, and Switchboard (\fig~\ref{fig:awe-swbd}), a conversational speech domain. We observe that the relative ranking of S3Ms differs for the two settings. For instance, \avhubert \ has better performance on Switchboard, outperforming all \baseM \ models, whereas all other S3Ms have higher scores on LibriSpeech. \wavlm-\largeM \ outperforms \wavlm-\baseM \ on Switchboard but the larger model under-performs on LibriSpeech. In both cases, the domain of pre-training data provides a potential explanation. Specifically, \avhubert \ models are pre-trained on TED videos~\cite{afouras2018lrs3} and \wavlm-\largeM \ is pre-trained on a mix of data~\cite{chen2021gigaspeech, wang2021voxpopuli} including orated speech and spontaneous speech, whereas all other S3Ms are trained on read speech domains~\cite{panayotov2015librispeech, kahn2020libri, hsu2020text}. 

We note that some cross-model rankings are consistent across evaluation domains. For instance, \hubert \ and \wavlm, both pre-trained to predict discrete cluster IDs from intermediate layers, outperform \wavtovec, which is trained to recover local features. As seen for other task-based evaluation (\sects~\ref{sec:word-seg},~\ref{sec:res-sts}), the visually grounded models, \fastvgsp \ and \vghubert, outperform the speech-only \baseM \ models, \wavtovec \ and \hubert, used to initialize them.

Additionally, we observe that the layer-wise trends for all S3Ms are consistent across evaluation domains and follow a similar dependence on the pre-training objective as noted by our previous results (\sect~\ref{sec:res-cca}) and some prior work~\cite{pasad2023comparative}.

\subsubsection{Word segmentation}
\label{sec:findings-domain-wseg}
We evaluate word segmentation on LibriSpeech (\fig~\ref{fig:wseg-libri}) and Buckeye (\fig~\ref{fig:wseg-buckeye}). Similarly to previous findings, we observe that the relative ranking of S3Ms differs for the two settings. Specifically, S3Ms pre-trained solely on LibriSpeech (\wavtovec-\baseM, \hubert-\baseM, \wavlm-\baseM) take a much larger hit in performance when evaluated on Buckeye, and the visually grounded models, on the other hand, have a slightly better performance on Buckeye than on LibriSpeech. 
Again, the layer-wise trends for most S3Ms are invariant to the evaluation domain. \wavlm-\largeM \ does not follow this trend and more than half of the layers have a drastically poorer performance on Buckeye. We hypothesize that the hyperparameters (tuned on LibriSpeech, \sect~\ref{sec:exp-wseg}) transfer better for other \largeM \ models than for \wavlm-\largeM, due to domain mismatch, as discussed above for {\it pool-AWD} (\sect~\ref{sec:findings-domain-awd}).

%% file: sections/conclusion.tex
\section{Conclusion}
The analyses presented here further our understanding of S3Ms, specifically their representation of word-level properties. Some of our findings corroborate patterns found in earlier work; for example, the most linguistically ``deep'' information appears to be encoded best in a small set of intermediate layers, and pre-training objective and model size influence layer-wise trends. We contribute new findings about previously unstudied aspects of S3Ms, such as the distribution of word information within word segments and the encoding of syntactic and semantic features. Most importantly, the comparison of a large number of models using the same analyses and tasks, and the study of multiple word-level properties, enables a more complete understanding of the space of S3Ms. As an additional product of this work, we obtained strong results on multiple benchmark tasks, outperforming prior work using simple models based on frozen S3M representations.

Our work studies which S3M layers are better (or worse) at encoding certain linguistic properties. Previous studies have used similar findings to guide modeling decisions when adapting pre-trained models for downstream tasks~\cite{pasad2023comparative, yang2023device, xie2022hidden}, including the choice of which layers to drop, distill\footnote{We use the term ``distill" to encompass various modeling variants such as transfer learning and using post-processed activations as targets, in addition to model distillation.}, or reinitialize~\cite{pasad2021layer, hwang2022pseudo, chang2022distilhubert, zaiem2023fine, Li2023parameterefficient, choi2021neural, hsu2021hubert}. We therefore expect our findings to inform design choices for both model development and their utilization for downstream tasks. For instance, for all the S3Ms we study, our analysis reveals that linguistic content is most prominent within the intermediate layers (\fig~\ref{fig:cca-scores}). Since layers encoding semantic content should be particularly beneficial for language understanding tasks, our findings suggest an exploration of alternative strategies to the common practice of adding a prediction head to the topmost layer~\cite{shon2022slue, shon2023slue}.

Our analyses have addressed several questions about S3Ms' word-level representations, thereby providing a foundation to address more challenging questions.
For example, a natural next step is to ask how much (and where) phrase- and sentence-level properties, such as constituents, dependencies, and entities, are encoded. For some tasks, such as word segmentation, although our results with S3Ms are stronger than prior work, they are far from solving the task.
Finally, we have noted (as have some prior studies) that larger models are not always better by all measures, raising the question of what the additional model capacity provides and whether there is a better way to train and utilize larger models.

%% file: sections/ack.tex
\section*{Acknowledgments}
We thank the anonymous reviewers and the action editor for their time and helpful feedback. This work is partially supported by AFOSR grant FA9550-18-1-0166.